\documentclass{article}

\usepackage[english]{babel}

\usepackage[letterpaper,top=2cm,bottom=2cm,left=3cm,right=3cm,marginparwidth=1.75cm]{geometry}

\usepackage{amsmath}
\usepackage{graphicx}
\usepackage[colorlinks=true, allcolors=blue]{hyperref}

\title{Navigating Fairness in AI Radiology: Concepts, Consequences, and Crucial Considerations}

\usepackage{authblk}

\author[1]{Vasantha Kumar Venugopal \thanks{Corresponding author: vasanthdrv@gmail.com}}
\author[1]{Abhishek Gupta \thanks{Corresponding author: abhishek.gupta840@gmail.com}}
\author[1]{Rohit Takhar}
\author[2]{Charlene Liew Jin Yee}
\author[3]{Catherine Jones}
\author[4]{Gilberto Szarf}
\affil[1]{CARPL.ai, New Delhi, India}
\affil[2]{Duke-NUS Medical School, Singapore}
\affil[3]{I-MED Radiology Network, Australia}
\affil[4]{Hospital Israelita Albert Einstein, Brazil}

\begin{document}
\maketitle

\begin{abstract}

Artificial Intelligence (AI) has significantly revolutionized radiology, promising improved patient outcomes and streamlined processes. However, it's critical to ensure the fairness of AI models to prevent stealthy bias and disparities from leading to unequal outcomes. This review discusses the concept of fairness in AI, focusing on bias auditing using the Aequitas toolkit, and its real-world implications in radiology, particularly in disease screening scenarios. 

Aequitas, an open-source bias audit toolkit, scrutinizes AI models' decisions, identifying hidden biases that may result in disparities across different demographic groups and imaging equipment brands. This toolkit operates on statistical theories, analyzing a large dataset to reveal a model's fairness. It excels in its versatility to handle various variables simultaneously, especially in a field as diverse as radiology. The review explicates essential fairness metrics: Equal and Proportional Parity, False Positive Rate Parity, False Discovery Rate Parity, False Negative Rate Parity, and False Omission Rate Parity. Each metric serves unique purposes and offers different insights. We present hypothetical scenarios to demonstrate their relevance in disease screening settings, and how disparities can lead to significant real-world impacts. 
\end{abstract}

\section{Key Points}

\begin{enumerate}
    \item The Aequitas toolkit is a powerful resource for auditing and mitigating bias, providing a wide range of metrics to help dissect and understand the potential biases lurking in AI models.
    \item Different fairness metrics serve distinct purposes in AI systems. In a disease screening scenario, metrics such as Equal and Proportional Parity, False Positive Rate Parity, False Discovery Rate Parity, False Negative Rate Parity, and False Omission Rate Parity play significant roles
    \item Understanding each metric's nuances is essential for ensuring a comprehensive approach to bias mitigation.
\end{enumerate}

\section{Introduction}

Artificial intelligence (AI) has staked its claim as a game-changer in healthcare, with an extraordinary potential to transform various aspects, including disease diagnosis, treatment, and management \cite{1-Ardila2019,2-Topol2019,3-Kooi2017}. Radiology, a field inherently dependent on image interpretation, has emerged as a prime candidate for the AI revolution \cite{4-Choy2018, 5-Shen2017}. Indeed, AI algorithms' ability to process and learn from vast volumes of imaging data offers an enticing prospect for enhancing radiologists' diagnostic accuracy and efficiency \cite{6-Ueda2021}. However, as we integrate AI deeper into radiology, it's essential to ensure these systems operate fairly, delivering consistent performance across diverse patient groups. Unfortunately, this isn't a given. 

AI models can unwittingly perpetuate or even amplify existing biases, leading to disparities in healthcare outcomes \cite{7-Obermeyer2019, 8-Rajkomar2018}. Such biases can originate from various sources, including the data used to train these models and the methods employed in their development [9]. It's a pressing concern that has garnered significant attention, prompting researchers to develop methods and tools to detect and mitigate AI biases \cite{8-Rajkomar2018, 9-Panch2019}. This review will delve into these critical areas, focusing on the concept of bias, disparity, and fairness, the Aequitas toolkit - a key player in assessing AI fairness, and the real-world impacts of these issues in radiology AI.

\section{Disparity, Bias, and Fairness – Definitions and Explanations}

Before diving into the mechanisms of bias in AI and its impact on radiology, it is essential to establish a common understanding of some critical terms: disparity, bias, and fairness. These terms, although sometimes used interchangeably, hold distinct meanings in the context of AI and healthcare.

Disparity refers to differences, variations, or inequalities that occur naturally or systematically within a population. In healthcare, these disparities can emerge across various dimensions, including age, gender, ethnicity, socioeconomic status, and more. Importantly, not all disparities are unjust or undesirable. For instance, the higher incidence of breast cancer in women compared to men is a biological disparity \cite{29-McCartney2019, 16-https://doi.org/10.48550/arxiv.1811.05577}. However, if certain ethnic groups receive less accurate cancer screenings due to algorithmic bias, this systematic disparity becomes a concern.

Bias, in the context of AI, refers to the systematic error introduced into the model that leads to unfair treatment or outcomes for certain groups. This could be due to several factors, such as skewed data representation, biased labels, or the algorithm's inherent limitations \cite{20-nelson2019bias}. This bias can be unintentional and subtle, yet it can lead to unfair or unequal outcomes \cite{21-SeyyedKalantari2021}.

Fairness, a broad and multifaceted concept, generally refers to the state where no group or individual is unduly disadvantaged or favored by an AI system \cite{14-https://doi.org/10.48550/arxiv.1810.08810}. In healthcare AI, fairness can be understood as the equal and impartial treatment of all individuals, regardless of their characteristics or group affiliations \cite{15-Dwork2012}. Achieving fairness is challenging, especially given the complex and multifaceted nature of healthcare data \cite{11-Zhang2022}. However, it's a pursuit worth striving for, considering the potential impacts of unfair AI systems on people's lives and health \cite{8-Rajkomar2018, 9-Panch2019}.

Understanding these concepts forms the foundation for exploring bias in AI, its detection, and its mitigation. In the following sections, we will delve into these subjects more deeply, focusing on the Aequitas toolkit and its relevance to radiology AI.

\section{Aequitas: A Bias and Fairness Audit Toolkit}
\subsection{Aequitas Toolkit - What is it?}
The Aequitas toolkit is an open-source bias audit tool, specifically created to bring transparency and fairness into the machine learning models used in various domains – healthcare, finance, criminal justice, and more. Developed by the Center for Data Science and Public Policy at the University of Chicago, Aequitas is a practical tool that invites us to take a closer look at the fairness of our AI systems\cite{16-https://doi.org/10.48550/arxiv.1811.05577}. The name Aequitas, derived from Latin, implies fairness and equality - and that's precisely what this toolkit is designed to promote \cite{16-https://doi.org/10.48550/arxiv.1811.05577}.

Aequitas is developed with the aim of making machine learning models more transparent, fair, and accountable \cite{16-https://doi.org/10.48550/arxiv.1811.05577}. It works by examining performance data from a machine learning model and generating an array of metrics that give insight into the model's fairness across various subgroups.

\subsection{Why Aequitas? What makes it the ideal tool for auditing bias in radiology AI?}

Firstly, it offers a comprehensive suite of fairness metrics, enabling a deep-dive analysis of the different facets of fairness in a model. The broad range of metrics allows for nuanced assessments of fairness, something that is crucial in healthcare \cite{19-pmlr-v136-oala20a}.

Secondly, Aequitas is designed to handle large and complex datasets, making it perfect for the vast volumes of data involved in radiology AI. It assumes the law of large numbers, meaning it works best when there's a significant sample size for each group you're comparing.

Lastly, Aequitas doesn't make assumptions about disease prevalence or incidence among different groups, an essential feature when considering fairness in disease screening scenarios. Instead, it looks at the model's predictions and outcomes across groups, allowing for a fair and unbiased comparison.

\section{Assessing Disparity and Parity}
While disparity refers to the differences or imbalances in the distribution of outcomes or predictions across different groups, its counterpart, parity refers to the fairness or equity in those distributions. 

\subsection{Disparity Quantification}
The aequitas bias toolkit employs disparate impact metrics for disparity calculation. As a cardinal metric in disparity computation, particularly in fairness analysis, disparate impact scrutinizes relative outcome or prediction distributions across heterogeneous groups and detects possible biases. The disparity is gauged relative to a reference group, given by:


\[Disparity Measure = \frac{Outcome Rate Metrics for specific group}{Outcome Rate Metrics for reference group}\]

The Outcome Rate Metrics could be any precision metrics such as False Negative Rate (FNR), False Omission Rate (FOR), Positive Predictive Value (PPV), etc. A Disparity Measure of 1 indicates no disparity, whereas values diverging from 1 suggest the existence of disparity.

\subsection{Reference Group Selection Strategy}
The reference group selection is a vital aspect of the aequitas bias toolkit analysis. It offers a standard against which other groups are juxtaposed to detect disparities and assess fairness. Several strategies can be deployed to select a reference group:
\begin{enumerate}
\item \textbf{Predominant Group as Reference:} The most widespread group within the dataset can serve as the reference group. It helps identify disparities or biases relative to the dominant group.

\item \textbf{Demographically Balanced Reference:} An averaged or balanced group, representing the demographic or protected attribute under analysis, can serve as the reference group to reflect the overall demographic constitution of the study population

\item \textbf{External Benchmark:} In specific instances, an external benchmark from existing research or datasets can be the reference group.

\item \textbf{Customized Reference:} Aequitas allows for reference group customization based on specific criteria or research objectives.
\end{enumerate}

The choice of the reference group should be aligned with the unique objectives and considerations of the medical imaging analysis.

\subsection{Parity}
Parity denotes fairness or equity across different demographic or protected groups. Parity achievement requires disparity minimization in the previously mentioned metrics. Aequitas employs a threshold, the disparity intolerance, representing the acceptable disparity level. When Disparity Measure values surpass this threshold, it indicates unacceptable disparity, implying the need for further inquiry and potential bias rectification. If the Disparity Measure lies within the disparity intolerance range, the group is deemed in parity with the reference group.


\begin{equation}
\label{eq:disparity}
\tau \leq \text{Disparity Measure} \leq \frac{1}{\tau}
\end{equation}

In equation \ref{eq:disparity}, $\tau$ represents the disparity intolerance, within the range [0,1]. By setting a specific limit, it offers a benchmark for evaluating disparity magnitude, aiding in distinguishing between tolerable disparities and those demanding intervention.

\section{Fairness Metrics – Their Relevance in Disease Screening Scenario}
Fairness metrics provide critical insights into the performance of AI models, particularly in the context of disease screening. They act as evaluative standards, ensuring that AI models are equitable across various patient demographics \cite{23-https://doi.org/10.48550/arxiv.1809.09245}.

\subsection{Equal and Proportional Parity}
These metrics ensure equal representation across diverse groups within a dataset.  These are crucial when you want to ensure representation among the groups in your dataset. These metrics ensure that all groups have equal chances of being flagged by the AI \cite{24-rieger2020explainability, 16-https://doi.org/10.48550/arxiv.1811.05577}. However, in a disease screening scenario, accurate disease identification is more important than equal representation. So, while these metrics are interesting, they may not be the main focus in a screening context.

Imagine we have a disease that is more prevalent in the older population. In this case, if the AI system flags older individuals more often for further testing, that is not necessarily a bias, but rather the algorithm correctly identifying higher-risk individuals. However, if the AI system disproportionately flags older individuals with a higher false positive rate, this could be seen as a bias against the older population, leading to unnecessary stress and potential harm from additional testing. Therefore, while it's important to consider equal representation, in a disease screening scenario, it is even more crucial to ensure accurate disease identification and avoid unnecessary harm. 

\subsection{False Positive Rate Parity}
This metric evaluates the proportion of healthy individuals incorrectly identified as having the disease within each demographic group \cite{25-raza2023auditing, 26-fairness-package}. A higher false positive rate in a particular group can lead to unnecessary additional testing, inducing potential harm and increased healthcare costs. FPR Parity ensures this proportion is the same across all groups \cite{16-https://doi.org/10.48550/arxiv.1811.05577}. 

This metric stipulates that each demographic group should experience the same rate of false-positive errors. For instance, if we consider ethnicity as a variable, false positive rate parity means that each ethnicity group—be it Asian, African American, or Caucasian—should encounter an identical rate of false positive errors. 

The significance of this metric comes into focus particularly when the consequences of false positives are severe and could lead to unnecessary interventions or adverse outcomes. Hence, ensuring false positive rate parity is crucial as it safeguards against disproportionately high false positive rates in any single group.

In the context of tuberculosis (TB) screening using an AI-powered radiology tool for immigration checks, the principle of False Positive Rate Parity is quite significant. This concept demands equal false positive rates across all demographic groups - for instance, various nationalities or ethnic groups undergoing the immigration check. 

Let's consider a scenario where the AI system is being used to screen chest X-rays of immigrants from various countries for signs of TB. A false positive in this context would mean that an individual who does not have TB is incorrectly identified by the AI system as having the disease. The consequence of this error could be a delay in their immigration process, or even a denial of immigration altogether. This could be seen as a punitive outcome, as it unjustly impacts the individual's prospects.

In this context, False Positive Rate Parity would mean that immigrants from all countries have an equal likelihood of experiencing a false positive. This ensures that no nationality or ethnic group is unduly affected by the errors of the AI system, maintaining fairness in the immigration process.

The False Positive Rate is calculated as:
\[FPR = \frac{FP}{FP + TN}\]
where:

FP stands for False Positives (the number of negative instances incorrectly classified as positive) \&
TN stands for True Negatives (the number of negative instances correctly classified as negative)	

\subsection{False Discovery Rate Parity}
This metric quantifies the proportion of false positives within the flagged cases for each group \cite{27-https://doi.org/10.48550/arxiv.1610.08452, 16-https://doi.org/10.48550/arxiv.1811.05577}. A higher false discovery rate indicates a larger share of the flagged cases in a group being false positives, leading to unnecessary medical interventions and patient anxiety.

This criterion asserts that every demographic group should experience an equivalent rate of false discovery errors. For example, when analyzing the factor of ethnicity, false discovery rate parity means all ethnic groups should have the same rate of false discovery errors.

The significance of this criterion similar to FPR comes to the fore especially when the outcome of a false discovery could lead to punitive consequences for individuals.

False Discovery Rate Parity is another critical principle in the same scenario of TB screening for immigration checks. This principle demands an equal rate of false discovery errors across all demographic groups. In this context, a false discovery would mean that among the individuals flagged by the AI system as having TB (i.e., the “discovered” cases), an equal proportion across all nationalities are actually false positives.

For instance, let's assume the AI system flags 100 immigrants from Country A and 50 immigrants from Country B as potential TB patients. If 30 out of the 100 flagged cases from Country A are false positives, then to maintain False Discovery Rate Parity, the number of false positives from the 100 flagged cases from Country B should also be around 15. This ensures that no nationality bears an undue burden of false discoveries, maintaining fairness in the immigration process.

The false discovery rate is the ratio of the number of false positive results to the number of total predicted positive test results.

\[FDR = \frac{FP}{PP}
      = \frac{FP}{FP+TP}\]

      where:
FP stands for False Positives (the number of negative instances incorrectly classified as positive),
PP stands for Predicted Positives (the number of positive predictions) \&
TP stands for True Positives (the number of positive instances correctly classified as positive)

\subsection{False Negative Rate Parity}
This metric is of particular importance in disease screening scenarios. A higher false negative rate indicates that the model overlooks more actual disease cases in one group than in others \cite{25-raza2023auditing, 26-fairness-package}. This bias could lead to delayed treatment and poorer health outcomes for these patients.

In the context of radiology AI, False Negative Rate Parity is a significant metric that calls for equal false negative rates across all identified groups \cite{19-pmlr-v136-oala20a}. Let's consider the example of gender groups in the patient population.

Visualize a scenario where an AI-driven radiology tool is deployed for lung cancer screening. A false negative in this context indicates that a patient who indeed has lung cancer is inaccurately identified by the AI system as cancer-free. This error can have serious implications as it may delay necessary treatment for the patient, possibly leading to detrimental health outcomes.

Let's take gender groups as an example - men, women, and non-binary individuals. If False Negative Rate Parity is achieved, it indicates that all these gender groups have an equal likelihood of encountering a false negative error. This is especially important when the role of the screening tool is to assist in early disease detection and subsequent provision of treatment.

In essence, if the AI system is intended to help identify patients who require medical intervention, such as lung cancer treatment, it becomes paramount to ensure that it's not missing patients from certain gender groups disproportionately. Adherence to the principle of False Negative Rate Parity can ensure that the AI system is equally effective across all gender groups, thereby promoting fairness and reducing health disparities.

\[FNR = \frac{FN}{FN + TP}\]

where:

FN stands for False Positives (the number of positive instances incorrectly classified as negative) \&
TP stands for True Positives (the number of positive instances correctly classified as positive)	

\subsection{False Omission Rate Parity}
False Omission Rate - FOR  is the fraction of false negatives of a group within the predicted negative of the group. It represents the proportion of actual positive cases within the predicted negative cases for each group \cite{16-https://doi.org/10.48550/arxiv.1811.05577}. A higher FOR suggests that a larger percentage of the non-flagged cases in a group are actual positive cases, indicating missed opportunities for early intervention \cite{19-pmlr-v136-oala20a} \cite{27-https://doi.org/10.48550/arxiv.1610.08452}.

Let's understand this with an example. Consider a radiology AI tool tasked with screening for osteoporosis, a condition that can occur across all age groups, although it's more common in older adults. In this context, a false omission error happens when the AI tool fails to flag an individual who actually has early signs of osteoporosis.

Let's say, for instance, a middle-aged individual, who is at the onset of osteoporosis, goes for a routine health checkup. The AI system, however, fails to flag the early signs, and the individual is told they don't need any further osteoporosis-related intervention. This is a false omission. Here, the person in question is actually in need of further examination and treatment but has been erroneously classified as not needing any.

Now, if we apply the concept of False Omission Rate Parity, it would mean that the AI system should have an equal probability of making this mistake across all age groups – young adults (18-35), middle-aged adults (36-60), and older adults (61 and above). This is crucial to prevent the AI tool from disproportionately missing disease cases in certain age groups, thus ensuring fairness in disease detection across all age groups.

By striving for False Omission Rate Parity, we can promote an equitable allocation of healthcare resources and can better ensure that early signs of diseases are not overlooked disproportionately in any age group.

\[FOR = \frac{FN}{PN}
      = \frac{FN}{FN+TN}\]

      where:
FN stands for False Negatives (the number of positive instances incorrectly classified as negative),
PN stands for Predicted Negative (the number of Negative predictions) \&
TN stands for True Negatives (the number of negative instances correctly classified as negative)

\section{Hypothetical Scenarios to Demonstrate the Impact of Bias in Screening Settings}
Let's imagine two hypothetical scenarios in medical screening settings where we'll observe the impact of bias. Here, we'll bring the concepts of False Positive Rate (FPR), False Discovery Rate (FDR), False Negative Rate (FNR), and False Omission Rate (FOR) to life.

\subsection{Scenario 1 to demonstrate the impact of FPR disparity: Tuberculosis Screening for Australian Visa Applications}
Australia's academic appeal attracts a diverse range of international students every year. In the 2021-22 period, the country received half a million student visa applications. The applicants came from all over the globe, but the majority hailed from five nations: China, India, the United Kingdom, the United States, and Vietnam.
\begin{figure}
\centering
\includegraphics[width=1\linewidth]{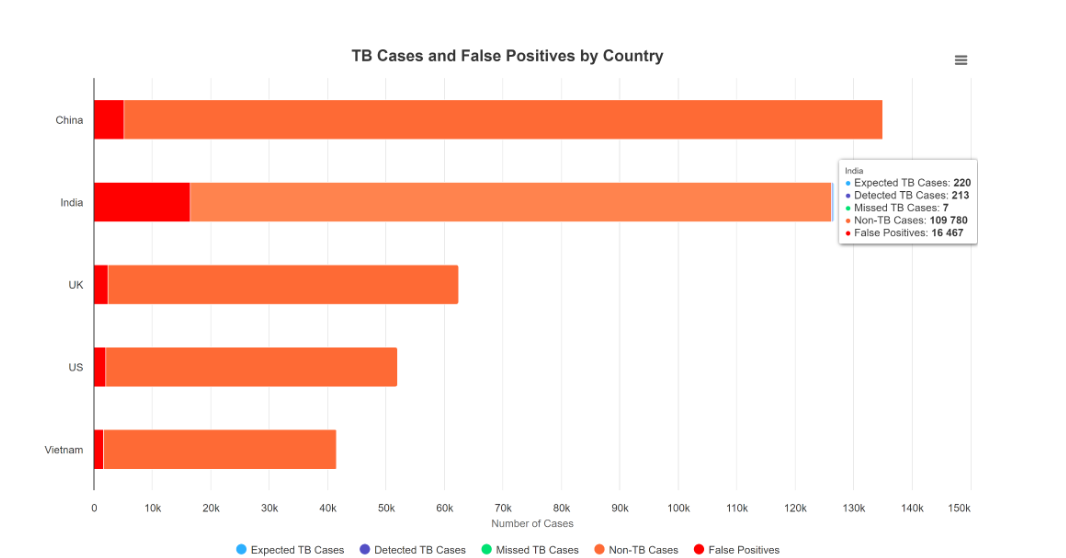}
\caption{\label{fig:Tb_cases}Hypothetical scenario to demonstrate the impact of the FPR disparity using a TB Immigration screening scenario. The assumed FPR disparity leads to a disproportionately huge penalty for the Indian immigrants}
\end{figure}
With 130,000 applications, Chinese students led the pack, making up 26\% of the total. Indian students followed closely with 110,000 applications, accounting for 22\% of all applications. The UK, the US, and Vietnam also made significant contributions with 60,000 (12\%), 50,000 (10\%), and 40,000 (8\%) applications, respectively \cite{17-who_tb_report} as shown in Figure \ref{fig:Tb_cases}.

Now, let's introduce our AI-powered tuberculosis screening tool into this immigration process. Let's also assume that the tool has a sensitivity of 97\% and a specificity of 96\% with a False Positive Rate (FPR) disparity towards Indian applicants, with a disparity ratio of 1.5. This bias can significantly affect the number of false positives among Indian applicants. 

To illustrate this, let's consider the TB prevalence in each of these countries. China's prevalence is 100 per 100,000 people, India's is double that at 200, the UK and the US have the lowest prevalence at 10 and 9 respectively, and Vietnam's prevalence mirrors China's at 100.

In the majority of cases, the tool performs well. For instance, in China, out of the 130 expected cases of TB, the tool correctly identifies 126 but misses 4. It also falsely identifies 5,200 healthy individuals as having TB. Similarly, in the UK and the US, the false positives are limited to 2,400 and 2,000, respectively. 

However, the picture changes when we turn to India. Here, the expected cases of TB rise to 220 in line with the high prevalence. The AI identifies 213 correctly but misses 7. The bias, though, results in a staggering 16,500 healthy individuals being flagged as having TB. That's more than triple the false positives in China, despite India having fewer applicants and only double the TB prevalence. 

This scenario highlights the substantial impact of bias in AI-powered disease screening tools. The bias against Indian applicants can lead to a significant increase in false positives, causing unnecessary stress and leading to a higher burden of additional testing. Such disparities underscore the importance of fairness in AI systems to prevent such disproportionate impacts.

\subsection{Scenario 2 to demonstrate the impact of FNR disparity: Lung Cancer Screening in Singapore}
Singapore also has a diverse population, with Malays comprising about 15\% of the residents. With the introduction of an AI tool for lung cancer screening, the hope was to ensure early detection and treatment across all ethnic groups. However, let's assume that our AI tool exhibits a False Negative disparity with a 1.6 disparity ratio towards the Malay population.

With 100,000 individuals undergoing screening, about 2,000 people are expected to have lung cancer given the 2\% prevalence rate observed in global studies. 
\begin{figure}
\centering
\includegraphics[width=1\linewidth]{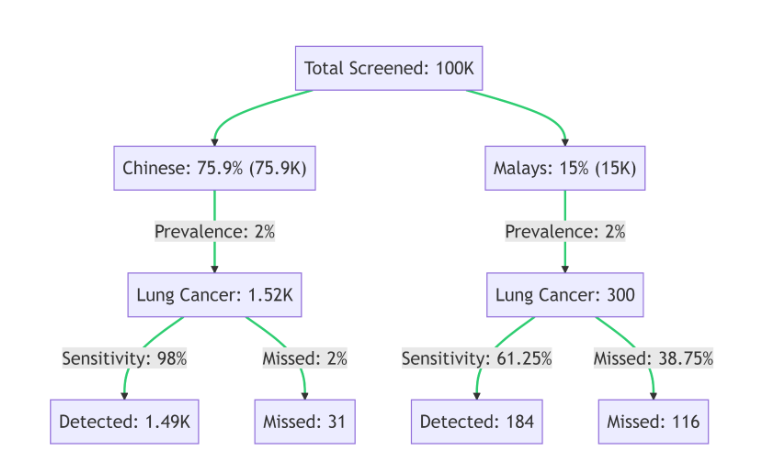}
\caption{\label{fig:LungCancertree}Flow chart describing the hypothetical FNR disparity scenario among the Singaporean population}
\end{figure}
Let's first consider the Chinese population, which represents 75.9\% of the total, or 75,900 individuals. With a disease prevalence of 2\%, about 1,518 Chinese individuals would have lung cancer as shown in Figure \ref{fig:LungCancertree}. Given the AI tool's sensitivity of 98\%, it would correctly identify about 1,487 of these cases, missing around 31 cases. \cite{18-homeaffairs_study_visa} 

\begin{figure}
\centering
\includegraphics[width=1\linewidth]{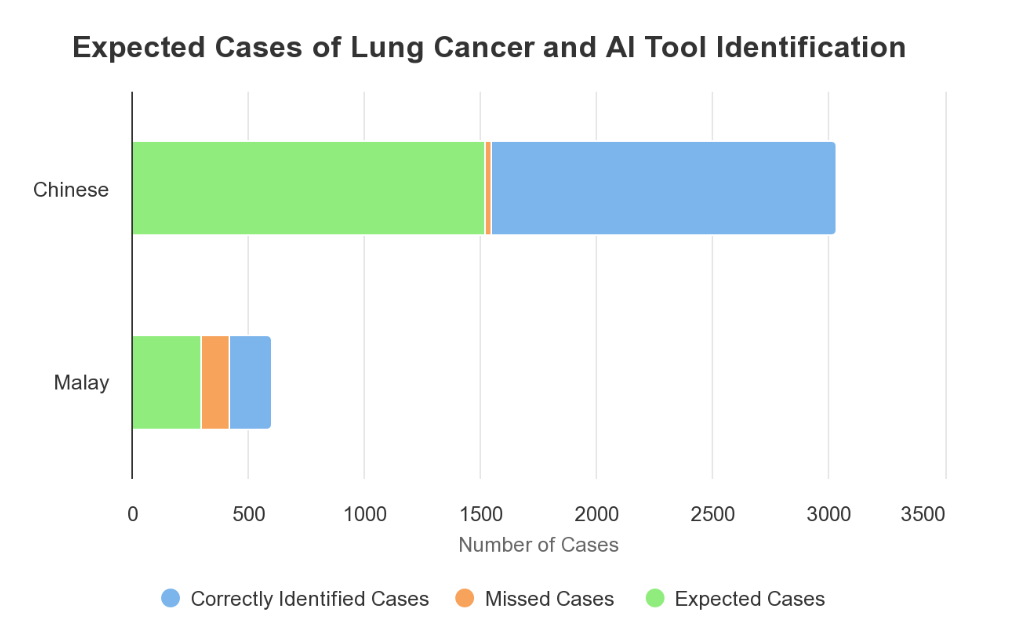}
\caption{\label{fig:lung_cancer}Hypothetical scenario to demonstrate the impact of the FNR disparity using a Lung cancer screening scenario. The assumed FNR disparity leads to a disproportionately high ratio of missed cancers among Malay patients}
\end{figure}

Now, let's consider the Malays, who represent 15\% of the total or 15,000 individuals. With a disease prevalence of 2\%, about 300 Malay individuals would have lung cancer. However, due to the AI tool's bias, the effective sensitivity for Malays drops to 61.25\% (98\%/1.6). This would mean the tool correctly identifies only about 184 of these cases, missing around 116 cases as in Figure \ref{fig:lung_cancer}.

Comparing the numbers, you can see the stark difference. Despite the Malay population being about a fifth of the Chinese population, the number of missed cases is nearly four times higher due to the bias in the AI tool. The impact is significant, demonstrating how bias can lead to disproportionate negative outcomes for certain groups. 

\section{Mitigation of bias}

While a detailed discussion of bias mitigation strategies is beyond the scope of this article, we want to present a broad overview of some recommended strategies. A multi-pronged strategy is required for mitigating bias in AI. Representative samples in training datasets must be ensured, AI learning algorithms should be adjusted for bias reduction, and post-training modifications to the model's decisions need to be made. Transparency and explainability in the AI decision-making process facilitate the pinpointing and correcting of biases. Regular audits, such as those enabled by Aequitas, serve as a vital tool for monitoring fairness over time.

\section{Conclusion}
Each of these metrics defined in the Aequitas package offers a unique perspective on fairness, highlighting potential biases in AI models\cite{16-https://doi.org/10.48550/arxiv.1811.05577, 22-https://doi.org/10.48550/arxiv.1810.01943}. In the context of disease screening, these metrics ensure that the models are not only accurate on average but are also fair across all groups \cite{12-Faghani2022}. The hypothetical scenarios presented herein elucidate the profound effects of compounded bias, underlining the importance of considering intersectionality in our audits. Our examination demonstrated that bias, left unchecked, could significantly skew diagnostic outcomes, potentially exacerbating existing health disparities.

Nevertheless, it is important to remember that fairness is context-specific[36]. The interpretation of these metrics must consider the specific characteristics of the disease, the population, and the overall screening strategy \cite{28-Mbakwe2023}. Therefore, the use of these metrics is a step towards a more equitable healthcare system, where all patients, irrespective of their demographics, receive accurate and timely disease detection \cite{10-Rouzrokh2022, 13-Kahn2022, 19-pmlr-v136-oala20a}.

\bibliographystyle{alpha}
\bibliography{reference}

\end{document}